\documentclass[11pt]{article} 
\pdfoutput=1
\usepackage{rldmsubmit,palatino}
\usepackage{algorithm}
\usepackage[noend]{algorithmic}
\usepackage{graphicx}

\title{Temporal-Difference Learning to Assist Human Decision Making during the Control of an Artificial Limb}

\author{
Ann L. Edwards\\
Department of Computing Science\\
University of Alberta\\
Edmonton, AB, Canada, T6G 2E8\\
\texttt{ann.edwards@ualberta.ca} \\
\And
Alexandra Kearney\\
Department of Computing Science\\
University of Alberta\\
Edmonton, AB, Canada, T6G 2E8\\
\texttt{kearney@ualberta.ca} \\
\And
Michael Rory Dawson\\
Glenrose Rehabilitation Hospital\\
Edmonton, AB, Canada, T5B 0B7\\
\texttt{mrd1@ualberta.ca} \\
\AND
Richard S. Sutton\\
Department of Computing Science\\
University of Alberta\\
Edmonton, AB, Canada, T6G 2E8\\
\texttt{rsutton@ualberta.ca} \\
\And
Patrick M. Pilarski\thanks{Please direct correspondence to Patrick M.\ Pilarski. All authors are affiliated with the Alberta Innovates Centre for Machine Learning (AICML) and the Reinforcement Learning \& Artificial Intelligence Laboratory (RLAI), University of Alberta. Pilarski is also affiliated with the Division of Physical Medicine \& Rehabilitation, Faculty of Medicine \& Dentistry, University of Alberta. \vfil Appears at: {\em The 1st Multidisciplinary Conference on Reinforcement Learning and Decision Making, Princeton, NJ, USA, Oct.\ 25--27, 2013.}}\\
Department of Computing Science\\
University of Alberta\\
Edmonton, AB, Canada, T6G 2E8\\
\texttt{pilarski@ualberta.ca} \\
}

\begin{document}

\maketitle

\begin{abstract}
In this work we explore the use of reinforcement learning (RL) to help with human decision making, combining state-of-the-art RL algorithms with an application to prosthetics. Managing human-machine interaction is a problem of considerable scope, and the simplification of human-robot interfaces is especially important in the domains of biomedical technology and rehabilitation medicine. For example, amputees who control artificial limbs are often required to quickly switch between a number of control actions or modes of operation in order to operate their devices. We suggest that by learning to anticipate (predict) a user's behaviour, artificial limbs could take on an active role in a human's control decisions so as to reduce the burden on their users. Recently, we showed that RL in the form of general value functions (GVFs) could be used to accurately detect a user's control intent prior to their explicit control choices. In the present work, we explore the use of temporal-difference learning and GVFs to predict when users will switch their control influence between the different motor functions of a robot arm. Experiments were performed using a multi-function robot arm that was controlled by muscle signals from a user's body (similar to conventional artificial limb control). Our approach was able to acquire and maintain forecasts about a user's switching decisions in real time. It also provides an intuitive and reward-free way for users to correct or reinforce the decisions made by the machine learning system. We expect that when a system is certain enough about its predictions, it can begin to take over switching decisions from the user to streamline control and potentially decrease the time and effort needed to complete tasks. This preliminary study therefore suggests a way to naturally integrate human- and machine-based decision making systems.
\end{abstract}


\acknowledgements{The authors gratefully acknowledge support from the Alberta Innovates Centre for Machine Learning, Alberta Innovates -- Technology Futures, and the Glenrose Rehabilitation Hospital Foundation. We also thank Thomas Degris and Jason P.\ Carey for a number of helpful discussions leading up to this work, and Adam Parker for his technical assistance and development on the exArm robotic system.}

\startmain 

\section{Introduction}

In this article we explore the use of reinforcement learning (RL) methods to assist in human decision making during the control of a human-robot interface.  We suggest that by acquiring and utilizing knowledge about a user's control-related decisions, control systems and human-machine interfaces could begin to take on an active role in human decision-making so as to reduce the burden on their users. Knowledge about a user and their robotic system can take the form of learned predictions about the interactions between the human and their device. 

Learning and maintaining a wide range of predictive sensorimotor knowledge has been demonstrated in recent work on Nexting (Modayil, White, and Sutton 2012) using learned General Value Functions (GVFs; Sutton et al. 2011). An extension of conventional RL value functions, GVFs represent temporally extended predictions about arbitrary signals of interest. GVFs can be learned in real time using standard RL methods, and have been successfully applied to gather anticipatory knowledge during ongoing human-robot interactions (Pilarski et al. 2012, 2013). As shown in our recent work, combining conventional control methods with GVF-derived predictions can potentially reduce the time and effort needed for users to control a switching-based human-machine interface (Pilarski et al. 2012; Pilarski and Sutton 2012).

Observations from motor control in the human brain also suggest that the ongoing prediction of motor control choices could potentially impact the intuitiveness and functionality of hybrid human-machine decision-making systems. There is a strong relationship between sensorimotor prediction and control in the human brain, and anticipated motor outcomes have been suggested as an important factor in generating and improving control  (Flanagan et al.\ 2003). As described by Flanagan et al.\ (2003) and Wolpert et al. (2001), predictions are thought to be learned by human subjects before they gain control competency. It is possible that similar mechanisms will prove beneficial for human-robot interaction (Fagg et al.\ 2004). In particular, leveraging learned knowledge stored in GVFs may be a viable way to support the control-related decisions made by a user with regard to their associated device.

\begin{figure*}[hb]
\centering
\includegraphics[width=0.73\linewidth]{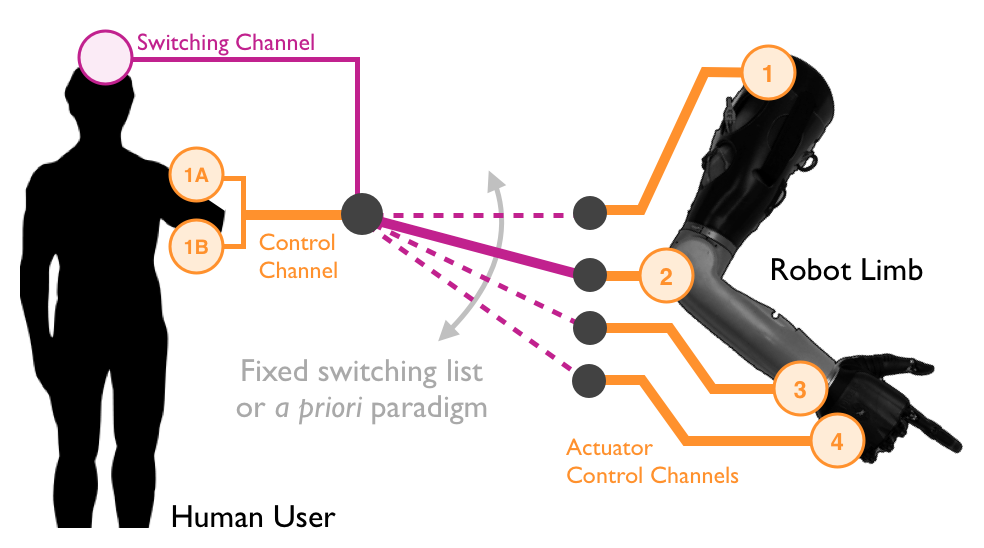}
\caption{{\em Example of function switching as used to control an assistive device}. One problem for human-machine interaction occurs when a machine's controllable dimensions outnumber the control channels available to its human user.}
\label{fig:switching}
\end{figure*}

As a motivating example, amputees who control artificial limbs are often required to quickly switch between a number of control actions or modes of operation in order to operate their devices. The increasing complexity of their component human-robot interfaces is in fact one of the major barriers to the use of modern artificial limbs. Artificial limbs commonly use recorded muscle signals (electromyographic recordings, or EMG) to actuate the different joints and  motors of a robot system. This approach is  termed {\em myoelectric control}. 
In more advanced myoelectric systems there are fewer EMG recording sites available on an amputee's body than there are degrees of freedom (DOF) in the prosthesis that the user must control (Williams 2011). One solution to this problem has been the use of EMG signals or mechanical toggles to enable a user to manually switch their control influence between the available joints, movements, grasping patterns, or functions available via the robot arm (Figure \ref{fig:switching}). While this approach has proved viable for functional use, it is often viewed by users as non-intuitive and unnatural, thereby increasing a user's cognitive effort and the time needed to complete a task. One notable example is the myoelectric interface for some commercial hand and forearm prostheses, which often require a sequence of muscle contractions and manual changes by the user to select a desired gripping or pinching pattern. As such, and despite the potential for restoring lost functions, many patients still reject the use of electromechanical artificial limbs (Williams 2011; Micera et al. 2010). 

In the present work, we therefore explore the unification of RL with conventional switching-based control interfaces. In particular, we demonstrate the use of online temporal-difference (TD) learning to predict when a user will switch their control influence between the different control functions of an articulated robot arm. We expect that when a system is certain enough about its predictions regarding a user's switching target and switching timing, it can begin to take over some function switching decisions from the user to streamline control and potentially decrease the time and effort needed to complete tasks. Our over-arching goal is to develop predictive approaches that ultimately enable the more natural control of complex assistive devices. 

\vspace{-0.6em}

\section{Methods}

\begin{figure*}[t]
\centering
\includegraphics[height=0.31\linewidth]{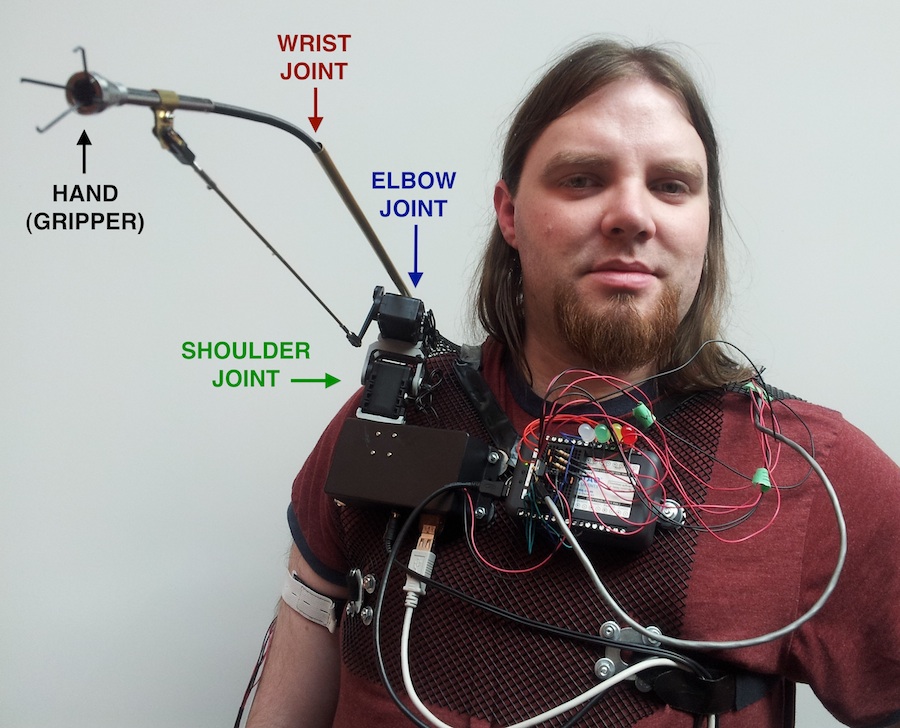} %
\includegraphics[height=0.31\linewidth]{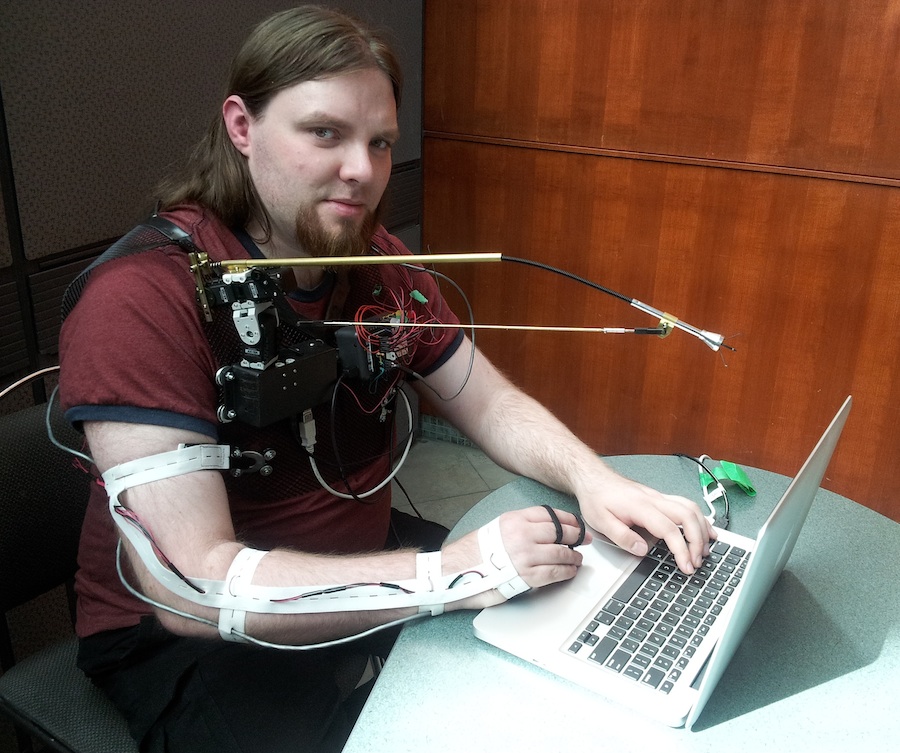}\\
\vspace{-0.4em}
\caption{{\em The experimental platform used in this work}: a wearable robot limb that is controlled using muscle signals from the human body, where the user sequentially controls and switches between the available joints using voluntary muscle contractions (similar to the control interface for a commercial forearm prosthesis).}
\label{fig:motion}
\end{figure*}

A wearable, myoelectrically controlled robot arm was used as the experimental platform for this work (Figure \ref{fig:motion}). This system had four controllable actuators. Two joints of the robot arm could be controlled to move the limb left, right, up, and down, approximating the motions provided by biological shoulder and elbow joints. The lower portion of the arm could also flex inward and outward as in wrist joint movement, and the arm terminated in a simple gripping actuator. Electrodes were affixed to the skin of non-amputee subjects and used to measure EMG signals from four different muscles on the user's body (DE-3.1 double differential electrodes and a Bangoli-8 acquisition system from Delsys, USA). These EMG signals were mapped to two control channels: one to actuate a robotic joint, and one to switch between the different joints in a fixed, sequential fashion.

\begin{figure}[t]
\centering
\includegraphics[width=0.49\linewidth]{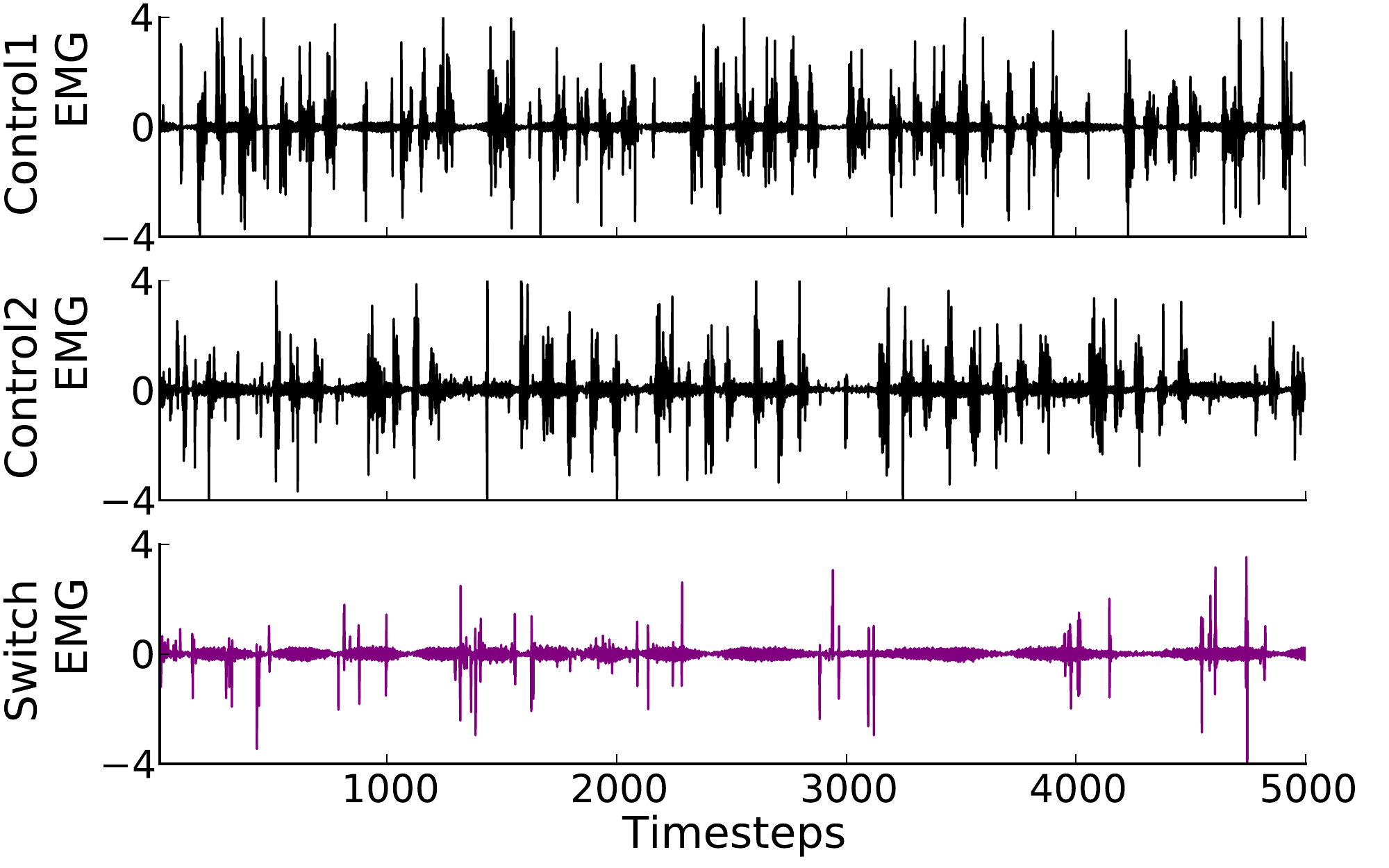} \hfil
\includegraphics[width=0.49\linewidth]{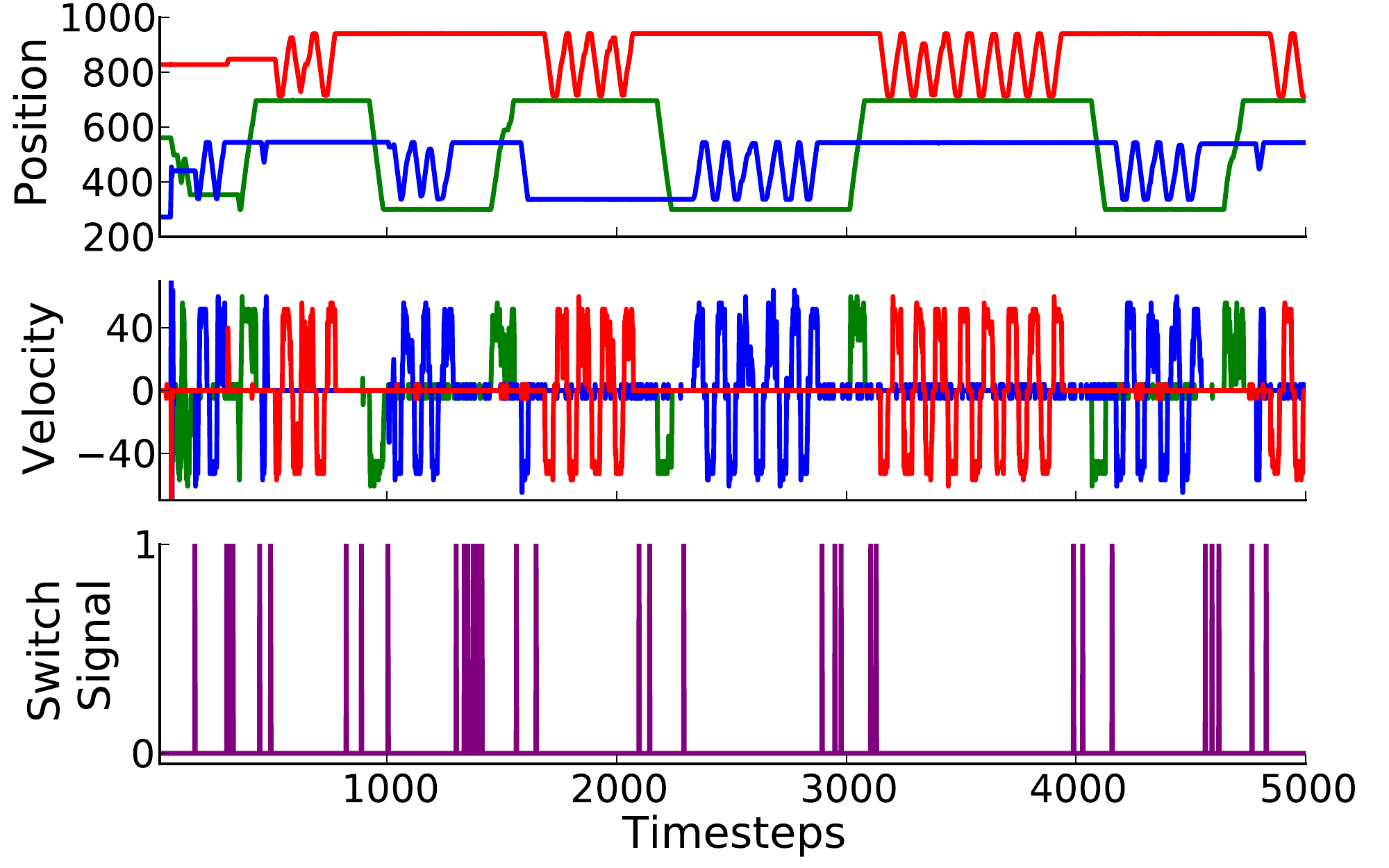}\\
\vspace{-0.7em}
\caption{{\em Example of the sensorimotor data stream from the human-machine interface}, including recorded muscle activity (EMG) for control and switching channels (black and purple traces, left), human switching actions (purple trace, right), and the angle and speed for three of the robot's joints (red, blue, and green traces, right).}
\label{fig:data}
\end{figure}

We examined the ability of GVF-based TD learning to predict joint switching from human interaction with the robot system during simple movement tasks. An able-bodied (non-amputee) subject actuated the myoelectric arm, using electrodes affixed over the wrist flexor and wrist extensor muscles of each arm. Using this wearable system, each subject performed a semi-repetitive motion, moving the robot's shoulder to the right, moving the elbow up and down an arbitrary number of times, moving the shoulder joint back to the left, and then moving the wrist up and down an arbitrary number of times. This H-shaped movement pattern was repeated for 10--30 minutes. As shown in Figure \ref{fig:data}, this resulted in a rich stream of data for use by the RL system, and provided a challenging setting for learning due to the temporal variability and non-stationary nature of the user's myoelectric control signals and switching behaviour.

The knowledge learned by our system regarding a user's switching actions took the form of temporally extended predictions about a user's switching prompts; these predictions were similar to the predictions made in our previous work on anticipating the activity of user-controlled actuators (Pilarski et al.\ 2012; Pilarski and Sutton 2012). Predictions were acquired and updated through multiple offline iterations using an implementation of GVFs (Sutton et al.\ 2011) and Nexting (Modayil, White, and Sutton 2012). Following the approach of Pilarski et al.\ (2012), GVFs were updated on each time step using TD learning with eligibility traces and tile-coding function approximation. Each GVF learner was initialized with parameters specifying the prediction of interest, including the timescale of the temporally extended prediction and the target signal of interest---here an on/off signal that was active when the user prompted the system to switch motor functions. Signal sampling and learning updates occurred at 15 Hz (many times per second). The state representation used by the machine learner was comprised of motor feedback (e.g., position, speed) from the robot arm and signals relating to the human's recorded muscle activity (e.g., processed EMG signals and switching cues), as well historical information in the form of decayed traces of these signals. As in previous work, our system also learned a series of temporally extended predictions regarding the motion of each user-driven joint and the user's myoelectric signals (with prediction done in the same way as for switching signal prediction, described above). 

\section{Results \& Discussion}

\begin{figure*}[ht]
\centering
\includegraphics[width=0.49\linewidth]{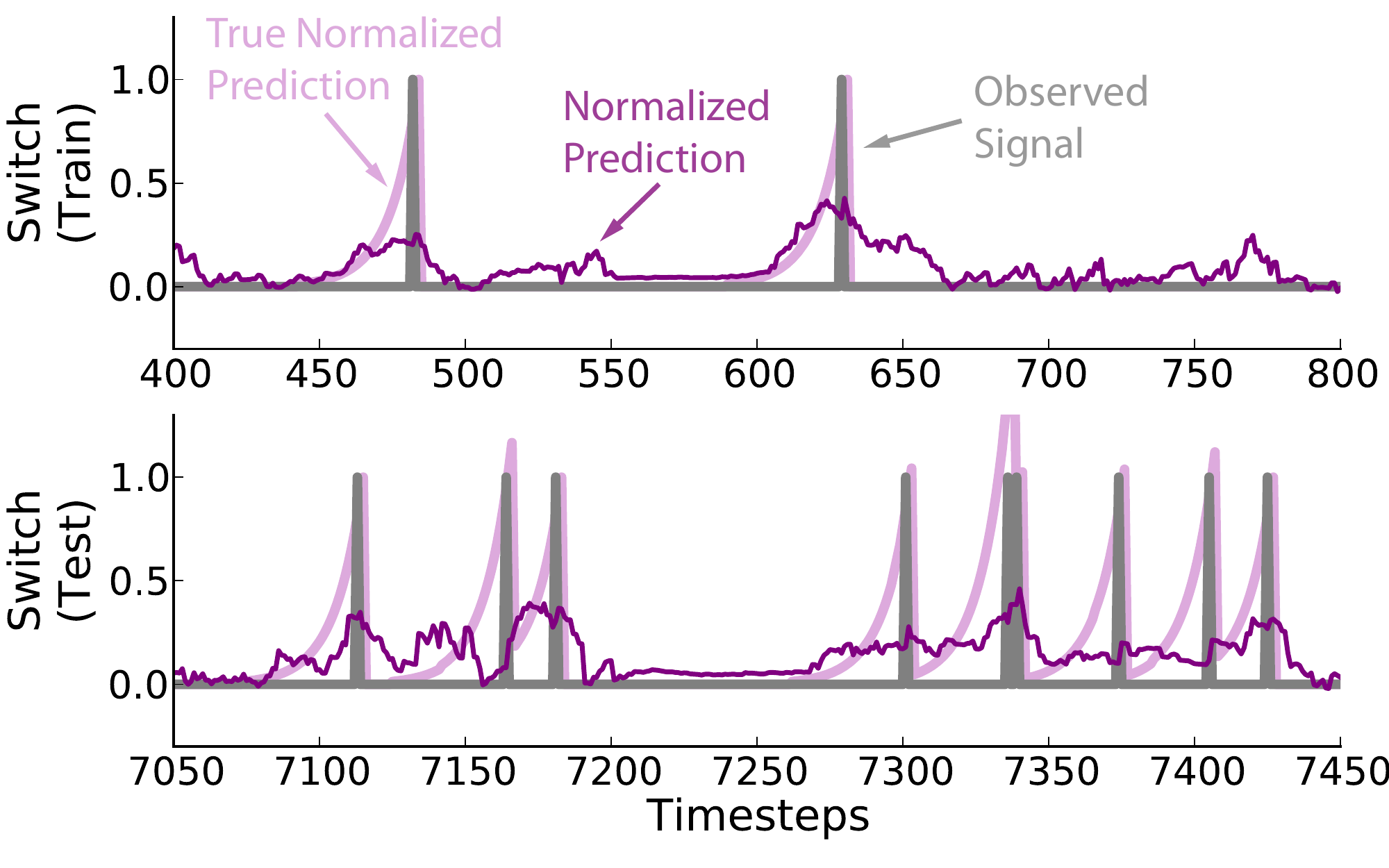} \hfil
\includegraphics[width=0.49\linewidth]{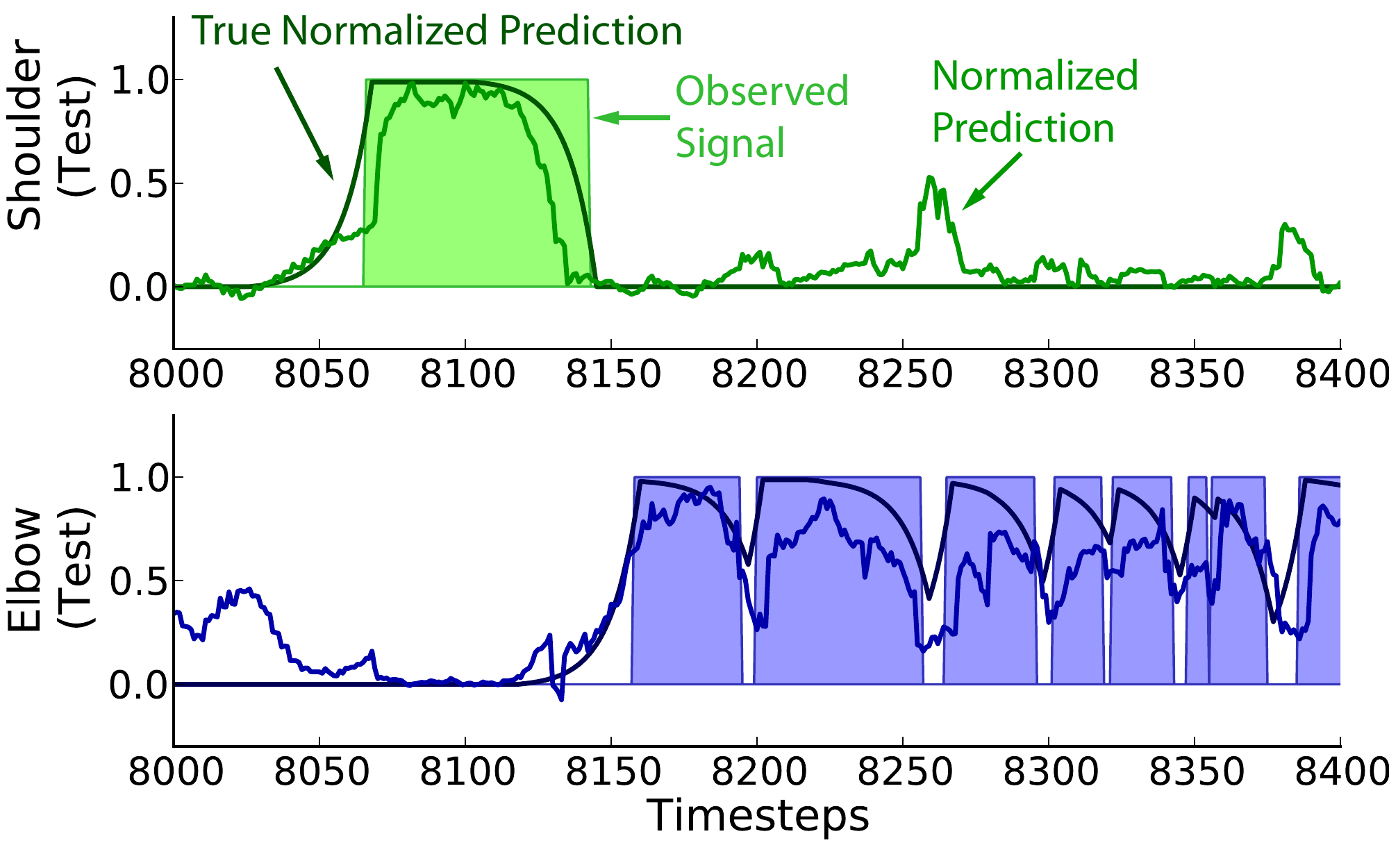}\\
\vspace{-0.7em}
\caption{{\em Example of switching event and joint activity predictions on previously unseen testing data} after five iterations of TD learning through 28k steps of training data. {\em Left}: switching event predictions begin to rise in advance of actual switching events initiated by the user, as shown on both training and testing data. {\em Right}: predictions about joint activity rise in advance of expected joint actuation. Modulating a control interface based on these predictions promises to reduce the time needed for a user to complete a task with a multi-joint robot arm (Pilarski et al., 2012, Pilarski and Sutton 2012). }
\label{fig:results}
\end{figure*}

As shown in Figure \ref{fig:results}, left, following a period of learning our approach demonstrated the ability to forecast an upcoming switching cue from the human user. Advance knowledge of switching (a rise in the dark purple trace) was observed to arrive a fraction of a second before the actual human-initiated event (grey pulse). Predictions on both training and testing data (dark purple traces) were also observed to begin to approach the true, computed return (light purple trace) as learned progressed. Additional learning is expected to improve the agreement between the true and learned predictions, and testing is ongoing to determine the best state representation for this learning scenario. As expected from previous work, our approach was also able to consistently anticipate which joint a human user would actuate next while performing their task (Figure \ref{fig:results}, right). The timescale for all predictions shown in Figure \ref{fig:results} was 10 time steps. Testing and training data were sampled from the same human user, with the training and testing sessions being conducted on different days. The testing data were not seen by the learning system prior to evaluation.

By ranking the magnitude of joint activity predictions prior to manual switching by the user, a learning system is able to determine the most appropriate joint to select at the time of switching (Pilarski et al. 2012). In other words, simple relationships between the predictions can be used to formulate the system's switching suggestions, i.e., which joint to actuate next. These suggestions depend on both context and learned knowledge about a user's preferences. The present work contributes a way to determine the desired timing of switching actions. Taken together, these straightforward applications of learned predictive knowledge provide a way to allow a learning-based control system to gradually assume more autonomy and decision-making responsibility during ongoing human-robot interactions. One useful feature of this approach is that no explicit or time consuming reinforcement is needed from the human user to correct or affirm the learning system's suggested decisions; the use of a mode or function by the user verifies the system's choices, while use of an alternate function decreases the learning system's predictions about the suitability of a given control option (Pilarski and Sutton 2012). Our approach therefore differs from predominant approaches to human-directed RL like human reward (e.g., Thomaz and Breazeal 2008) and demonstration learning (e.g., Lin 1992). 

 The ability shown in the present work to anticipate a switching event promises to greatly reduce the need to manually initiate switching. The time needed for switching could potentially even be eliminated in certain situations. As suggested in Pilarski and Sutton (2012), removing the need to explicitly switch functions during commonly performed tasks could result in almost as great a time savings as selecting the optimal switching target or function. These expectations remain to be demonstrated in future work with non-amputee and amputee subjects.

\section{Conclusions}

In this work we demonstrated the use of reinforcement learning (RL) to help with human decision making, and specifically provided a first step towards intuitive human interaction with a switching-based biomedical robot. Function switching is a common way to deal with increasing device complexity, but it poses additional challenges to the natural and efficient control devices by a user. To help address the barriers to streamlined human-robot interactions, we deployed state-of-the-art RL techniques to acquire and maintain knowledge about a user and their robotic system. Our approach was able to build up and maintain forecasts about a user's switching behaviour in real time. We also confirmed previous observations that our approach can detect a user's control intent prior to their explicit control actions. Bringing together these two ideas, a system could potentially determine what function a user intends to deploy, and when they wish to begin using the new function. Our approach allows a user to remain in direct control of a system while still allowing the device to suggest or initiate increasingly appropriate control options. Furthermore, the opportunity for ongoing yet optional human interaction in the decision-making process provides an intuitive and reward-free way for users to correct or reinforce the decisions made by a semi-autonomous machine learning system. This preliminary study therefore opens the way for naturally blending the control decisions made a human and their assistive robot or other human-machine interface. Future work will continue to pursue the integration of biological and synthetic reinforcement learning and decision-making systems.

\section*{References}

\parindent=0pt
\small
\def\hangin{\hangindent=0.15in}
\parskip=5pt

\hangin
 \noindent Fagg, A. H., Rosenstein, M. T., Platt, Jr., R., Grupen, R. A. (2004). Extracting user intent in mixed initiative teleoperator control. In \textit{Proceedings of the AIAA 1st Intelligent Systems Technical Conference}, Chicago, Illinois, 2004-6309.

\hangin
 \noindent Flanagan, J.~R., Vetter, P., Johansson, R.~S., Wolpert, D.~M. (2003). Prediction precedes control in motor learning. \emph{Current Biology} 13(2): 146--150.

\hangin
 \noindent Lin, L.-J. (1992). Self-improving reactive agents based on reinforcement learning, planning and teaching. \textit{Machine Learning} 8: 293--321.

\hangin
 \noindent Micera, S., Carpaneto, J., Raspopovic, S.\ (2010). Control of hand prostheses using peripheral information. \textit{IEEE Reviews in Biomedical Engineering} 3: 48--68.

\hangin
 \noindent Modayil, J., White, A., Sutton, R.~S. (2012). Multi-timescale nexting in a reinforcement learning robot. In \textit{Proceedings of the 2012 Conference on Simulation of Adaptive Behavior}, Odense, Denmark, 299--309. arXiv:1112.1133 [cs.LG]

\hangin
 \noindent Pilarski, P.~M., Dawson, M.~R., Degris, T., Carey, J.~P., Sutton, R.~S. (2012). Dynamic switching and real-time machine learning for improved human control of assistive biomedical robots. In \textit{Proceedings of the 4th IEEE International Conference on Biomedical Robotics and Biomechatronics (BioRob)}, June 24--27, Roma, Italy, 296--302. 
 
\hangin
 \noindent Pilarski, P.~M., Sutton, R.~S. (2012). Between instruction and reward: Human-prompted switching. \textit{AAAI 2012 Fall Symposium on Robots Learning Interactively from Human Teachers (RLIHT)}, Nov. 2-4, Arlington, VA, USA, AAAI Technical Report FS-12-07, 46--52.

\hangin
 \noindent Pilarski, P.~M., Dawson, M.~R., Degris, T., Carey, J.~P., Chan, K.~M., Hebert, J.~S., Sutton, R.~S.\ (2013). Adaptive artificial limbs: A real-time approach to prediction and anticipation. \textit{IEEE Robotics \&  Automation Magazine} 20(1): 53--64.

\hangin
 \noindent Sutton, R.~S., Modayil, J., Delp, M., Degris, T., Pilarski, P.~M., White, A.,  Precup, D. (2011). Horde: a scalable real-time architecture for learning knowledge from unsupervised sensorimotor interaction. In \textit{Proceedings of 10th International Conference on Autonomous Agents and Multiagent Systems (AAMAS)}, May 2--6, Taipei, Taiwan, 761--768.

\hangin
\noindent Thomaz, A.~L., Breazeal, C. (2008). Teachable robots: understanding human teaching behaviour to build more effective robot learners. \textit{Artificial Intelligence} 172: 716--737.

\hangin
 \noindent Williams, T.~W.\ (2011). Guest editorial: progress on stabilizing and controlling powered upper-limb prostheses. \textit{Journal of Rehabilitation Research and Development} 48(6): ix--xix.
 
\hangin
 \noindent Wolpert, D.~M., Ghahramani, Z., Flanagan, J.~R. (2001). Perspectives and problems in motor learning. \emph{Trends Cogn.\ Sci.} 5(11): 487--494.

\end{document}